\documentclass[a4paper, 12pt]{extarticle}
\usepackage{geometry}
\usepackage{subfig}

\newcommand{\KL}[2]{\mathrm{KL}\left( #1 \| #2 \right)}

\newcommand{\argmax}[1]{\underset{#1}{\arg \ \max \ }}

\newcommand{\indep}{\perp \!\!\! \perp}

\newcommand{\Phat}{{\hat{\mathcal{P}}}}

\newcommand{\Pa}{{\mathcal{P}}}
\newcommand{\Pdkk}{\mathrm{Part}_d^k}
\newcommand{\Sdk}{\mathrm{Set}_d^k}
\newcommand{\Pdk}{\mathrm{Part}_d^k}
\newcommand{\Sd}{\mathrm{Set}_d}
\newcommand{\Pd}{\mathrm{Part}_d}

\usepackage{amsfonts}
\usepackage{amsmath}
\usepackage{amssymb}
\usepackage[boxed]{algorithm2e}
\usepackage{hyperref}
\usepackage{cleveref}
\usepackage{graphicx}
\usepackage{color}
\usepackage{float}
\usepackage{caption}
\usepackage{diagbox}

\SetKwInOut{Input}{input}
\SetKwInOut{Output}{output}

\providecommand{\keywords}[1]
{
  \textbf{\textit{Keywords---}} #1
}

\newcommand{\bigzero}{\mbox{\normalfont\Large\bfseries 0}}
\newcommand{\rvline}{\hspace*{-\arraycolsep}\vline\hspace*{-\arraycolsep}}

\crefname{example}{example}{examples}
\crefname{theorem}{theorem}{Theorems}
\crefname{lemma}{lemma}{Lemmas}
\crefname{proposition}{proposition}{Propositions}
\crefname{remark}{remark}{Remarks}
\crefname{corollary}{corollary}{Corollaries}
\crefname{definition}{definition}{Definitions}
\crefname{equation}{equation}{Equations}
\crefname{table}{table}{Tables}
\crefname{figure}{figure}{Figures}
\crefname{section}{section}{Sections}

\begin{document}

\title{ISDE: Independence Structure Density Estimation}
\date{}
\author{Louis Pujol\thanks{Université Paris-Saclay, CNRS, Inria, Laboratoire de Mathématiques d’Orsay, 91405, Orsay, France. louis.pujol@universite-paris-saclay.fr}\\ }
\maketitle

\begin{abstract}
  In this paper, we propose ISDE (Independence Structure Density Estimation), an algorithm designed to estimate a multivariate density under Kullback-Leibler loss and the Independence Structure (IS) model. IS tackles the curse of dimensionality by separating features into independent groups. We explain the construction of ISDE and present some experiments to show its performance on synthetic and real-world data. Performance is measured quantitatively by comparing empirical $\log$-likelihood with other density estimation methods and qualitatively by analyzing outputted partitions of variables. We also provide information about complexity and running time.
\end{abstract}


\newpage
\pagenumbering{arabic}

\section{NOTATIONS}\label{sec:notations}

Let $f$ be a density function (a nonnegative real function whose integral is equal to $1$) over $\mathbb{R}^d$. If we think of $f$ from a statistical viewpoint, it is natural to refer to the indices $\{1, \dots, d\}$ as the features.

Let $S \subset \left\{1, \dots, d\right\}$, we denote by $f_S$ the marginal density of $f$ over $S$. For all $x=(x_1, \dots, x_d) \in \mathbb{R}^d$
 \begin{equation}
    f_S(x) = \int f(x) \prod_{i \notin S} dx_i.
\end{equation}
With a slight abuse of notation, to highlight the fact that $f_S(x)$ does not depend on $(x_i)_{i \notin S}$, we write $f_S(x_S)$ instead of $f_S(x)$.

Let $k$ be an positive integer not greater than $d$. We denote by $\Sdk$ the set of all subsets of $\{1, \dots, d\}$ with cardinal not greater than $k$ and by $\Pdk$ the collection of all partitions of $\{1, \dots, d\}$ constructed with blocks in $\Sdk$. We also use the shortcuts $\Sd = \Sd^d$ and $\Pd = \Pd^d$.

\section{INTRODUCTION}\label{sec:intro}

\paragraph{Unsupervised Learning and Density Estimation} Unsupervised learning is an important field of data analysis. It aims to design methods to extract meaningful information from a dataset with little prior knowledge. A central task in unsupervised learning is density estimation. Given a sample $X_1, \dots, X_N$ drawn independently from a random variable $X$ on $\mathbb{R}^d$ with a density $f$, the goal is to build an estimator $\hat{f}$ of $f$. This question finds many applications, and density estimation is a building block for many learning tasks such as clustering (\cite{chazal2013persistence}, \cite{campello2013density}) or anomaly detection (\cite{chandola2009anomaly}) among others.

\paragraph{Nonparametric and Parametric Density Estimation} The easiest way to do density estimation is to consider parametric models: data is supposed to be drawn from a probability distribution known up to a finite-dimensional parameter $\theta$. Estimating the density is then equivalent to estimating $\theta$. One example is the centered multivariate Gaussian framework, where the parameter $\theta$ is the covariance matrix $\Sigma$. An introduction to parametric statistics can be found in \cite{wasserman2004all}, chapter 9. This approach suffers from a lack of flexibility as it strongly constrains the model. At the other end of the spectrum lies nonparametric density estimation. In this framework, densities are no longer considered members of some finite-dimensional family but are supposed to belong to a set of functions with a given regularity. An introduction to the subject can be found in \cite{tsybakov2004introduction}.

\paragraph{Kernel Density Estimators} In the sequel, we focus on nonparametric density estimation. Kernel Density Estimator (KDE) is a popular density estimator in this context. It has its origins in the works of Rosenblatt \cite{Rosen1956} and Parzen  \cite{parzen1962estimation}. It has been successfully used to real-world applications in recent years (connectivity among salmon farms \cite{cantrell2018use}, physical activity \cite{king2015use}, ecological niche modelling \cite{qiao2017cautionary}, modelling of T cell receptors \cite{owens2009t}, among many others).

In this paper, we will consider Spherical Gaussian KDE (SGKDE). For a given bandwidth $h>0$ we define the SGKDE associated to $h$ and to the sample $X_1, \dots, X_N$ as 
\begin{equation}
    \hat{f}_h(x) = \frac{1}{N} \sum_{i=1}^N \frac{\exp\left( - \frac{\left( X_i - x \right)^{\mathrm{T}}\left( X_i - x \right) }{2h^2} \right)}{(2 \pi)^{d/2} h^d}.
    \label{Gaussiankde}
\end{equation}

As we will not consider other choices of kernels, we write KDE instead of SGKDE. The construction of the estimator over a data sample corresponds to the choice of the bandwidth. Different approaches exist. In practice, a cross-validation scheme over a collection of potential values of $h$ is a popular choice. See \cite{van2004asymptotic} for analysis in the context of maximum likelihood density estimation.

\paragraph{Curse of Dimensionality} When dealing with multidimensional data, one must be aware of the issues that the number of features can imply. It is a general fact that for the majority of statistical tasks, the higher the dimension is, the harder the estimation is (see, for example, \cite{giraud2021introduction}). For density estimation, the complexity can be evaluated through minimax risk, quantifying the statistical error in a worst-case scenario. It is influenced by two parameters: a regularity parameter $\beta$ and dimension $d$ , the rate of convergence for the squared $L_2$ loss is typically proportional to $N^{\frac{-2 \beta}{2 \beta + d}}$ (see \cite{goldenshluger2012adaptive} for a review of the literature). We remark that the higher $d$ is, the slower the minimax risk tends to zero. This phenomenon is a manifestation of the so-called curse of dimensionality. For practitioners, it should be adventurous to use a multivariate density estimator if the sample size is limited and the dimension becomes large, especially in the case of nonparametric estimation. A solution is to assume that unknown density belongs to a class of structured functions.

\paragraph{Moderately High Dimension Setting} In recent years,  attention was put on high-dimensional problems, where the number of features can vary from hundreds to thousands. We are interested here in situations of moderately high dimension, where the number of features can vary from a few ones to a few dozens. In this setting, the curse of dimensionality still occurs. It is of particular interest to distinguish both paradigms as we will develop algorithmic solutions that allow exhaustive search over admissible structures in moderately high dimensions but become too time-consuming in high dimensions.

\paragraph{Structural Density Estimation with Undirected Graphical Models} A way to consider a structure for a multivariate random variable is through its undirected graphical model (introduction to the field can be found in  \cite{giraud2021introduction} and more in-depth cover in \cite{wainwright2008graphical}). As we will not consider directed graphical models, we always consider that graphs are undirected in the sequel. Given a graph $G=(V, E)$ whose vertices correspond to the features $\{1, \dots, d\}$ we say that $G$ is a graphical model for $X$ if the following condition is satisfied:
\begin{equation}
     (i, j) \notin E \Rightarrow X^i \indep X^j | (X^k)_{k \notin (i, j)}. 
\end{equation}

Constraints on the graphical model associated with a distribution impose a structure on the density, and such a structure can help overcome the curse of dimensionality. However, learning a graphical model is a complex task in many situations. The general result is that if $G$  is a graphical model for a $d$-dimensional random variable $X$, denoting by $\mathcal{C}$ the set of cliques of $G$ (\textit{ie} fully connected sets of nodes), it exists a collection of nonnegative functions $(\psi_C)_{C \in \mathcal{C}}$ such that the density $f$ of $X$ can be written as
\begin{equation}
    f(x) = \frac{1}{Z} \prod_{C \in \mathcal{C}} \psi_C(x_c)
\end{equation}
where $Z$ is a normalization constant. As remarked in section 2.1.2 of \cite{wainwright2008graphical}, the functions $\psi_C$ do not have a clear relationship with the marginal densities of $f$. The density estimation under a graphical model for general graphs is then too ambitious, and it is necessary to constrain the graph structure.

\paragraph{Forest Density Estimation} In a fully nonparametric setting, to our knowledge, one method is available: Forest Density Estimation (FDE) \cite{liu2011forest}. It corresponds to the estimation of a density with an uncyclic graphical model (also called a forest). In this case, the density can be expressed with 1 and 2-dimensional marginals. If $G = (V, E)$ is a forest, the density $f$ of a random variable admitting $G$ as a graphical model enjoys the following formulation
\begin{equation}
    f(x) = \prod_{(i, j) \in E} \frac{f_{\{i, j\}}(x_i, x_j)}{f_{\{i\}}(x_i)f_{\{j\}}(x_j)} \prod_{k=1}^d f_{\{k\}}(x_k).
\end{equation}

In \cite{liu2011forest} the algorithm to estimate a forest and the corresponding density is presented. Let us emphasis that it requires the estimation of marginals up to dimension 2. Theorem 9 in \cite{liu2011forest} emphasis that if the true density enjoys a forest graphical model and under suitable condition on the density, the speed of convergence of FDE under Kullback-Leibler (KL) loss is related to the the speed of convergence for KDE in dimension $2$ instead of in the ambient dimension $d$. This emphasize that FDE is a remedy to the curse of dimensionality. The KL loss between $f$ and an estimator $\hat{f}$ is defined as
\begin{equation}
    \KL{f}{\hat{f}} = \int \log\left( \frac{f}{\hat{f}} \right)f. 
\end{equation}

\paragraph{Independence Structure} In the present work, we focus on the model of Independence Structure (IS) for multivariate density developed by \cite{lepski2013multivariate} and studied by \cite{rebelles2015mathbb}. It contains $d$-dimensional densities, which can be decomposed as a product of low-dimensional marginals, forming a partition of the original features.
\begin{equation}
    f(x) = \prod_{S \in \mathcal{P}} f_S(x_S)
\end{equation}

Under a graphical model perspective, it corresponds to graphs that are composed of disjoint connected components. Previous works on IS have highlighted that if the density enjoys the property that the size of the biggest block of the partition is equal to $k<d$, then the complexity of density estimation, measured through minimax rate of convergence under $L_p$ losses ($1\leq p\leq \infty$) is related to $k$ instead of the ambient dimension $d$. However, these works rely on the analysis of estimators that are hardly implementable for reasonable data size.

\paragraph{Our Contribution} We present Independence Structure Density Estimation (ISDE), a method designed to simultaneously compute a partition of the features and a density estimation as a product of marginals over this partition in order to maximize the empirical $\log$-likelihood, or equivalently, minimize the KL loss. Our method enjoys reasonable running time for moderately high-dimensional problems and can be combined with any density estimation technique, so it covers parametric and nonparametric settings. To our knowledge, we are the first to design an algorithm estimating as IS in the context of KDE.

\paragraph{Organization of the Paper} In \cref{sec:algorithm} we present the construction of ISDE. We compare our method with some existing ones for density estimation for synthetic datasets in \cref{sec:synthetic} and for real-world datasets in \cref{sec:realdata} before analyzing its algorithmic complexity and running time in \cref{sec:runningtime}.

\section{ISDE}\label{sec:algorithm}
This section presents ISDE, an algorithm designed to simultaneously perform density estimation and independence partition selection in a moderately high-dimensional setting. 



\paragraph{Specifications} Let $k$ be an input parameter. We aim to provide a method taking point cloud as input and outputting an IS (a partition of the features in $\Pdkk$) and a density estimator as a product of marginal estimators 
\begin{equation}
    \hat{f}_{\hat{\Pa}, \hat{h}_{\hat{\Pa}}} = \prod_{S \in \hat{\Pa}} \hat{f}_{S, \hat{h}_S}
\end{equation}
where $\hat{h}_{\hat{\Pa}} = \left( \hat{h}_S \right)_{S \in \hat{\Pa}}$ is a list of bandwidths. For $S \in \Sdk$, $\hat{f}_{S, h_S}$ denotes an estimator of the form \ref{Gaussiankde} constructed with the features belonging to $S$.

\paragraph{Number of Partitions vs. Number of Subsets} Before starting the explanation of how ISDE works, let us highlights some comparison between the number of partitions in $\Pdk$ and the number of subsets in $\Sdk$.

Let us start by comparing $S_d$ and $B_d$, the respective cardinals of $\Sd$ and $\Pd$. We have $S_d = 2^d - 1$ and $B_d$ is known as the Bell number of order $d$. \cref{npartnset} shows how these quantities compare for dimension lying between $10$ and $15$.

\begin{table}[H]
\centering
\resizebox{360pt}{!}{%
\begin{tabular}{|c|c|c|c|c|c|c|}
\hline
d & $10$ & $11$ & $12$ & $13$ & $14$ & $15$ \\
\hline
\hline
$S_d$ & $1,023$ & $2,047$ & $4,095$ & $8,191$ & $16,383$ & $32,767$ \\ 
\hline
$B_d$ & $115,975$ & $678,570$ &  $4,213,597$ & $27,644,437$ &  $190,899,322$ & $1,382,958,545$ \\
\hline
\end{tabular}%
}
\caption{Number of partitions vs number of subsets} \label{npartnset}
\end{table}

We remark that the number of partitions is much higher than the number of features. Even if we restrict ourselves to small values of $k$, the difference remains important. We denote $S_d^k$ and $B_d^k$ the cardinals of $\Sdk$ and $\Pdk$. It is simple to see that
\begin{equation}
    S_d^k = \sum_{i=1}^k \binom{d}{k}.
\end{equation}
For $B_d^k$ exact computation is harder but we can prove that (see \cref{computB2d})
\begin{equation}
    B_d^k \geq B_d^2 = 1 + \binom{d}{2} + \frac{\binom{d}{2}\binom{d-2}{2}}{2!} + \frac{\binom{d}{2}\binom{d-2}{2}\binom{d-4}{2}}{3!} \dots + \frac{\binom{d}{2} \dots \binom{d - 2\left(\lfloor d/2 \rfloor - 1\right)}{2}}{(\lfloor d / 2\rfloor )!}
\end{equation}

and notice that $B_d^2 \underset{d \rightarrow \infty}{\sim} d^\frac{d}{2} $ while $ S_d^k \underset{d \rightarrow \infty}{\sim} d^k $. For values of $d$ corresponding to moderately high-dimensional settings, some computations are gathered in \cref{npartnsethighdim} (the values of $B_d^2$ are approximations).

\begin{table}[H]
\centering
\begin{tabular}{|c|c|c|c|c|}
\hline
d & $20$ & $30$ & $40$ & $50$  \\
\hline
\hline
$S_d^3$ & $1,350$ & $4,525$ & $10,700$ & $20,875$ \\ 
\hline
$B_d^2$ & $2.4 \times 10^{10}$ & $6.1\times 10^{17}$ & $7.3 \times 10^{25}$ & $2.8 \times 10^{34}$ \\ 
\hline
\end{tabular}%
\caption{Number of partitions vs number of subsets} \label{npartnsethighdim}
\end{table}

These computations indicate that it would be beneficial to find a way to avoid the computation of $B_d^k$ estimators. Intuitively, as estimators are combinations of marginals estimators, it seems reasonable to decouple marginal estimations from partition selection. We will now see that we must carefully choose the loss function to implement this idea.

\paragraph{Choice of Loss Function} We have announced in the introduction that ISDE aims to minimize the Kullback-Leibler loss between the proper density and the estimate one. Here we will see that this choice is not innocuous and that other choices of loss function do not lead to a feasible algorithm.

In density estimation literature, the most popular choice for the loss function is undoubtedly the squared $L_2$ loss. For a partition $\mathcal{P} \in \Pdk$ we want to find the collection of bandwidth $(\hat{h}^\Pa_S)_{S \in \Pdk}$ solutions of
\begin{equation}
    \min_{(\hat{h}^\Pa_S)_{S \in \Pa}} \int \left( f - \hat{f}_{\Pa, h_\Pa} \right)^2 = \int \hat{f}_{\Pa, h_\Pa}^2 - 2 \int \hat{f}_{\Pa, h_\Pa} f + \int f^2.
\end{equation}
If $P[.]$ corresponds to the integral over the measure induced by the density $f$, an equivalent formulation is given by
\begin{equation}
    \min_{(h^\Pa_S)_{S \in \Pa}} \int \hat{f}_{\Pa, h_\Pa} ^2 - 2 P\left[ \hat{f}_{\Pa, h_\Pa}  \right] = \min_{(h^\Pa_S)_{S \in \Pa}}  \prod_{S \in \Pa} \int \hat{f}_{S, h_S}^2 - 2 P\left[ \prod_{S \in \Pa} \hat{f}_{S, h_S} \right].
\end{equation}

Let $S \in \Sdk$ and $\Pa_1, \Pa_2 \in \Pdk$ such that $S \in \Pa_1$ and $S \in \Pa_2$. There is no reason to have $\hat{h}_S^{\Pa_1} = \hat{h}_S^{\Pa_2}$ from the previous formulation. Then under the squared $L_2$ loss we have no clue on how we can avoid constructing as many estimators as elements in $\Pdk$. 

Now, for the KL loss, we want to find a collection of bandwidth $(\hat{f}_S)_{S \in \Pdk}$ minimizing
\begin{equation}
    \min_{(h^\Pa_S)_{S \in \Pa}} \int \log\left( \frac{f}{\hat{f}_{\Pa, h_\Pa}} \right) f.
\end{equation}
An equivalent formulation is given by
\begin{equation}
    \max_{(h^\Pa_S)_{S \in \Pa}} P \left[ \log  \hat{f}_{\Pa, h_\Pa} \right] = \max_{(h^\Pa_S)_{S \in \Pa}} \sum_{S \in \Pa} \left\{ P \left[ \log \hat{f}_{S, h_S} \right] \right\}
\end{equation}
using the property that the logarithm changes products into sums and the linearity of the operator $P[.]$. By opposition of what we have seen for the squared $L_2$ loss, if $S \in \Pa_1$ and $S \in \Pa_2$, we will have $h_S^{\Pa_1} = h_S^{\Pa_2}$. Then under KL loss, bandwidths optimization over marginal estimators and partition selection can be decoupled, leading to the necessity of computing $S_d^k$ density estimators instead of $B_d^k$. As shown in \cref{npartnset} and \cref{npartnsethighdim}, it leads to an appreciable gain in terms of algorithmic complexity.
    
\paragraph{Empirical Formulation of the Optimization Problem} Under KL loss, bandwidths optimization and partition selection become two separated tasks. This decoupling incites us to design an algorithm consisting of two steps: first, compute a marginal estimator $\hat{f}_S$ for all $S \in \Sdk$ and then find the best combination of them for a $\log$-likelihood criterion. Let $n$ and $m$ be two positive integers such that $m + n = N$. The dataset $X_1, \dots, X_N$ is split into two disjoint subsamples:
\begin{itemize}
    \item $W_1, \dots, W_m$ used to compute marginal estimators $(\hat{f}_S)_{S \in \Sdk}$
    \item $Z_1, \dots, Z_n$ used to compute empirical $\log$-likelihoods $(\ell_n(S))_{S \in \Sdk}$ where $\ell_n(S) = \frac{1}{n} \sum_{i=1}^n \log\left( \hat{f}_S(Z_i) \right)$
\end{itemize}

Let us use the notation $\ell_n(\Pa) = \sum_{S \in \Pa}\ \ell_n(S)$. The empirical optimization task can be written as

\begin{equation}\label{eq:optiempiric}
    \max_{\Pa \in \Pdk} \ell_n(\Pa)  = \max_{\Pa \in \Pdk} \sum_{S \in \Pa}\ \ell_n(S).
\end{equation}

\paragraph{Partition Selection} A naive approach to solve \ref{eq:optiempiric} is to compute $\ell_n(\Pa)$ for every partition of $\Pdk$ and then find the optimal one. However, this approach becomes time-consuming when $d$ grows and infeasible for large values of $d$ because of the number of partitions. Therefore, it will be appreciable to reformulate this optimization to speed up computation. It is possible to reformulate \ref{eq:optiempiric} as the following linear programming task.

Solve
\begin{equation}
    \max_{x \in \mathbb{R}^{\Sdk}} \sum_{S \in \Sdk} \ell_n(S) x(S)
\end{equation}

Under constraints
\begin{align}
    Ax &= (1, \dots, 1)^\mathrm{T} \\
    x &\in \{0, 1\}^{S_d^k}.
\end{align}
Where $x$ is a binary vector representing which elements of $\Sdk$ are selected, and $A$ is a $d \times S_d^k$ matrix where each column is a binary vector representing the composition of one of the sets of $\Sdk$. The condition $Ax = (1, \dots, 1)^\mathrm{T}$ then ensures that each feature is chosen once, implying that the sets selected with $x$ form a partition.

We validate this approach through a running time comparison (see \cref{comptimelpnaive}) between the implementation of a brute-force approach and a linear program solver. In this experiment, we fix the quantities $(\ell_n(S))_{S \in \Sdk}$, the brute-force approach consists in a for loop (implemented in Python), computing $\ell_n(\Pa)$ for all $\Pa \in \Pdk$ and returning the maximum. For the LP formulation, the optimization is done with the branch-and-bound method, implemented in the Python package PulP \cite{Mitchell11pulp:a}. With the brute-force approach and choice $k=d$, partition selection takes approximately 3 hours in dimension $15$ but less than $10$ seconds with LP formulation.

\begin{table}[H]
\centering
\label{comptimelpnaive}
\begin{tabular}{|c|c|c|c|c|c|c|c|}
\hline
d & $9$ & $10$ & $11$ & $12$ & $13$ & $14$ & $15$ \\
\hline
\hline
Brute-Force Approach & $0.2$ & $0.9$ & $5.2$ & $32$ & $219$ & $1304$ & $10437$ \\ 
\hline
LP Solver & $0.1$ & $0.2$ & $0.4$ & $0.8$ & $1.9$  & $4.1$ & $9.1$    \\
\hline
\end{tabular}%
\caption{Running time (seconds): linear programming vs brute-force approach for partition selection}
\end{table}

\paragraph{Conclusion} The resulting algorithm is \cref{algo}. It enjoys the following properties:

\begin{itemize}
    \item It exploits the decoupling of marginal density estimation and partition selection offered by choice of KL as discrepancy measure: it optimizes over partitions in $\Pdk$ even if it only requires the computation of $\Sdk$ marginal estimators.
    \item It is versatile: even if we present the construction of ISDE using KDEs for marginal estimation, it is possible to use any other base multivariate density estimator.
\end{itemize}

\begin{algorithm}[ht]
 \Input{$X_1,\dots,  X_N \in \mathbb{R}^d$, $k$ integer with $k \leq d$, integers $m$ and $n$ and a subroutine to perform multidimensional density estimation}
 \Output{Partition $\Phat \in \Pdk$, marginal estimates $(\hat{f}_S)_{S \in \Phat}$}
 \Begin{
 \For{$S \in \Sdk$}{
    Compute $\hat{f}_S(W_1, \dots, W_m)$ thanks to the density estimation subroutine
    
    Compute $\ell_n(S)$
 }
 Compute $\Phat \in \argmax{\Pa \in \Pdk} \sum_{S \in \Pa}\ \ell_n(S)$ using linear programming formulation}
\caption{ISDE}
\label{algo}
\end{algorithm}

\section{EXPERIMENTS ON SYNTHETIC DATA}\label{sec:synthetic}

In this section, we validate the performance of ISDE on synthetic data generated under IS hypothesis.

\paragraph{Data Generating Process} For a given list of positive integer (a structure) $S = [s_1, \dots, s_K]$, the data generating process is defined as follows. For each $s_i \in S$, we define a $s_i$ dimensional dataset drawn from $P_i$:
\begin{itemize}
    \item If $s_i = 1$, $P_i$ is the uniform distribution over $[0, 1]$
    \item If $s_i = 2$, $P_i$ is a distribution corresponding to data sample near two concentric circles with different radii
    \item If $s_i = 3$, a sample $X$ from $P_i$ is obtained as follows: let $Y_1$ and $Y_2$ be two independent Bernoulli variables with probability of success $0.5$ and $Y_3 = |Y_1 - Y_2|$.  $X$ is then drawn from the multivariate Gaussian distribution $\mathcal{N} \left( (Y_1, Y_2, Y_3), 0.08 \times I_3 \right)$. This is a situation where features of $P_i$ are pairwise independent but not mutually independent
    \item  If $s_i \geq 4$, $P_i$ is a mixture of two multivariate Gaussian distributions, one centered in $(0, \dots, 0)$, the other in $(1, \dots, 1)$
\end{itemize}
The final dataset results from their concatenation, plus featurewise rescaling so that each value lies between $0$ and $1$. The dimension is $d=\sum_{i=1}^k s_i$. This rescaling step does not affect the IS as it is done featurewise.

\paragraph{Evaluation Scheme} To evaluate the performance of an estimator, we compute the empirical $\log$-likelihood on a validation set $X^{\text{valid}} = X^{\text{valid}}_1, \dots, X^{\text{valid}}_M$ drawn independently from the same distribution as $X_1, \dots, X_N$:
\begin{equation}
    \text{Score}(\hat{f}) = \frac{1}{M} \sum_{i=1}^M \log\left( \hat{f}\left( X^{\text{valid}}_i \right) \right).
\end{equation}

The set $X^{\text{valid}} = X^{\text{valid}}_1, \dots, X^{\text{valid}}_M$ is not used to tune the estimators. In the experiments of this section, we set $M = 5000$.

\paragraph{Benchmarked Methods}

We will compare three density estimation algorithms for samples corresponding to different structures.

The first one is CVDKE, a KDE estimator where the bandwidth parameter is selected through a $5$-fold cross-validation to maximize empirical $\log$-likelihood on test data. The collection of possible bandwidths is a regular grid on a $\log$-scale from $0.01$ to $1$ with $30$ values.

The second one is ISDE with $k=d$ (\textit{ie} all partitions are tested), $m=n=0.5N$ and the collection of marginal estimators $(\hat{f}_S)_{S \in \Sd}$ is a collection of CVKDE estimators constructed with the sample $W_1, \dots, W_m$.

The third one is FDE. Our implementation is a slight modification of the held-out data approach proposed in \cite{liu2011forest}: we rely on the quantities $(\ell_n(S))_{S \in \mathrm{Set}_d^2}$ computed in ISDE as estimators of the quantities $\left( \int \log(f_S) f_S \right)_{S \in \mathrm{Set}_d^2}$. We use a cross-validation scheme to optimize the bandwidth instead of the plug-in approach presented in the paper.

We insist that comparing these methods for density estimation through empirical $\log$-likelihood for validation data is fair as all of them aim to maximize the $\log$-likelihood.

\paragraph{Results} Empirical $\log$-likelihood on validation data for methods listed above are shown in \cref{Lossnonparam}, for different structures and for the choice $N=5000$. Each experiment is repeated $5$ times, and we show the mean $\log$-likelihood and the standard deviation on the table.

\begin{table}[H]
\centering
\begin{tabular}{|l|c|c|c|}
\hline
 & [2, 2, 1] & [3, 3, 3] & [4, 4, 2, 2] \\
\hline
\hline
ISDE & $\mathbf{1.83 \pm 0.08}$ & $\mathbf{4.05 \pm 0.15}$ & $\mathbf{6.30 \pm 0.25}$\\
\hline
FDE & $\mathbf{1.83 \pm 0.08}$ & $2.88 \pm 0.14$ & $5.89 \pm 0.33$ \\
\hline
CVKDE & $0.56 \pm 0.03$ & $3.49 \pm 0.11$ & $3.96 \pm 0.16$\\
\hline
\end{tabular}%
\caption{Empirical $\log$-likelihood on validation data for different density estimators} \label{Lossnonparam}
\end{table}

\paragraph{Conclusion} For $[2, 2, 1]$, ISDE and FDE give similar results as they output the same graph and the same bandwidths. They both outperform CVKDE. For $[3, 3, 3]$, as features are pairwise independent, FDE outputs at every try a graph without any edge and computes the density as a product of one-dimensional marginals, leading to poor results in comparison to ISDE. CVKDE leads to better estimation for this setting than FDE but is outperformed by ISDE. For $[4, 4, 2, 2]$, FDE outputs a subgraph of the actual graphical model at every try. It leads to better estimation than CVKDE but worse than ISDE, which learns the proper IS at every try.

Thus, ISDE leads to better results than FDE and CVKDE for the task of structured density estimation under KL loss under IS. We interpret the bad performance of CVKDE as a manifestation of the curse of dimensionality. ISDE outperforms FDE because it considers potential higher-order dependencies between features than FDE, which only considers pairwise associations. However, let us remark that FDE covers some models not addressed by ISDE. ISDE performs better on data where IS is true, but we recommend testing both methods to determine the one that best fits the data.

We also remark that ISDE recovers exactly the IS for the considered settings. One can wonder why we do not observe that outputted partitions are not precisely the IS but partitions where blocks are a union of blocks of the true IS. We believe that this is because a useless merging of blocks in the partition is strongly penalized by ISDE as the dimension limits our ability to estimate a density accurately. Then the hold-out scheme implemented in ISDE (by splitting $X$ into $W$ and $Z$ in \cref{algo}) penalizes sufficiently too big blocks in partitions and leads to accurate recovery of IS.

\section{EXPERIMENTS ON MASS CYTOMETRY DATA}\label{sec:realdata}

This section is devoted to the presentation of some outputs on real-world datasets. In addition to studying the performance of ISDE in terms of $\log$-likelihood, it is the occasion to illustrate how we can interpret the outputted partition.

\paragraph{Datasets} The datasets presented here are the output of mass cytometry experiments. Cytometry allows high-throughput measurements at a single-cell level over a cell sample. Two types of information about cells are collected. Some are about the cell's geometry, and others about the abundance of some targeted proteins at their surface. The number of events for cytometry experiments on blood samples usually lies between $10,000$ and $1,000,000$, and the number of features can vary from a few ones to approximately $50$.

We present here results on two public cytometry datasets used in a benchmark of clustering methods paper \cite{weber2016comparison}, Levine13 and Levine32. Both are experiments on bone marrow cells extracted from healthy human donors with respectively 13 and 32 features. The preprocessing step is a featurewise rescaling to force each feature to take values in $[0, 1]$.

\subsection{Quantitative evaluation}

\paragraph{Benchmarked Algorithms} As in the previous section, we compare FDE, CVKDE, and ISDE (the value of $k$ depends on the dimension, we selected $k=3$ for Levine32 and $k=5$ for Levine13 to keep computations fast). 

We have also added a parametric approach to the benchmark: a Gaussian Mixture (GM) model with a selection of the number of components. This model is particularly adapted to cytometry as we naturally expect in this context that the data forms clusters representing cell populations (\cite{reiter2016clustering}, \cite{finak2009merging}).

Let $n_C$ be a positive integer corresponding to the number of components in the mixture. Let $p = \left( p_1, \dots, p_{n_C} \right)$ be a collection of nonnegative real number such that $\sum_{i=1}^{n_C} p_i = 1$, $\mu = \left( \mu_1, \dots, \mu_{n_C}\right)$ a collection of vector in $\mathbb{R}^d$ and $\Sigma = \left( \Sigma_1, \dots, \Sigma_{n_C}\right)$ a collection of $d \times d$ definite positive matrices. The density $f_{(n_C, p, \mu, \Sigma)}$ of the Gaussian mixture model associated with the parameters $(n_C, p, \mu, \Sigma)$ is
\begin{equation}
    f_{(n_C, p, \mu, \Sigma)} = \sum_{i=1}^{n_C} p_i f_{\mu_i, \Sigma_i}
\end{equation}
where $f_{\mu_i, \Sigma_i}$ is the density of the multivariate Gaussian random variable with mean $\mu_i$ and covariance matrix $\Sigma_i$.

Given $n_C$ and a dataset, it is possible to compute estimators $(\hat{p}, \hat{\mu}, \hat{\Sigma})$ with the EM algorithm \cite{dempster1977maximum} to maximize the $\log$-likelihood. As we do not know the optimal number of components in advance, a strategy is to fit a Gaussian mixture model for different $n_C$ (from $1$ to $30$ in our experiments) and select the number of components in the mixture with a cross-validation scheme. We rely on the implementation of these methods provided by scikit-learn \cite{sklearn_api} with no restriction on the shape of the covariance matrices.

Though GM is principally used for clustering purposes, it can also be interpreted as a parametric density estimator intended to maximize the $\log$-likelihood. It is then relevant to compare it with the other introduced methods.

\paragraph{Experimental Setup} From each dataset we have extracted a train sample with $N=5000$ events, this train sample is exclusively used to compute estimators $\hat{f}_{\text{CVKDE}}$, $\hat{f}_{\text{FDE}}$, $\hat{f}_{\text{ISDE}}$ and $\hat{f}_{\text{GM}}$. For ISDE we fixed $m=3000$ and $n=2000$. Then to compare between these density estimators, we sampled $20$ datasets with $2000$ events from the data that were not used to compute estimators.

\begin{figure}[H]%
    \centering
    \subfloat[\centering Levine13]{{\includegraphics[width=5cm]{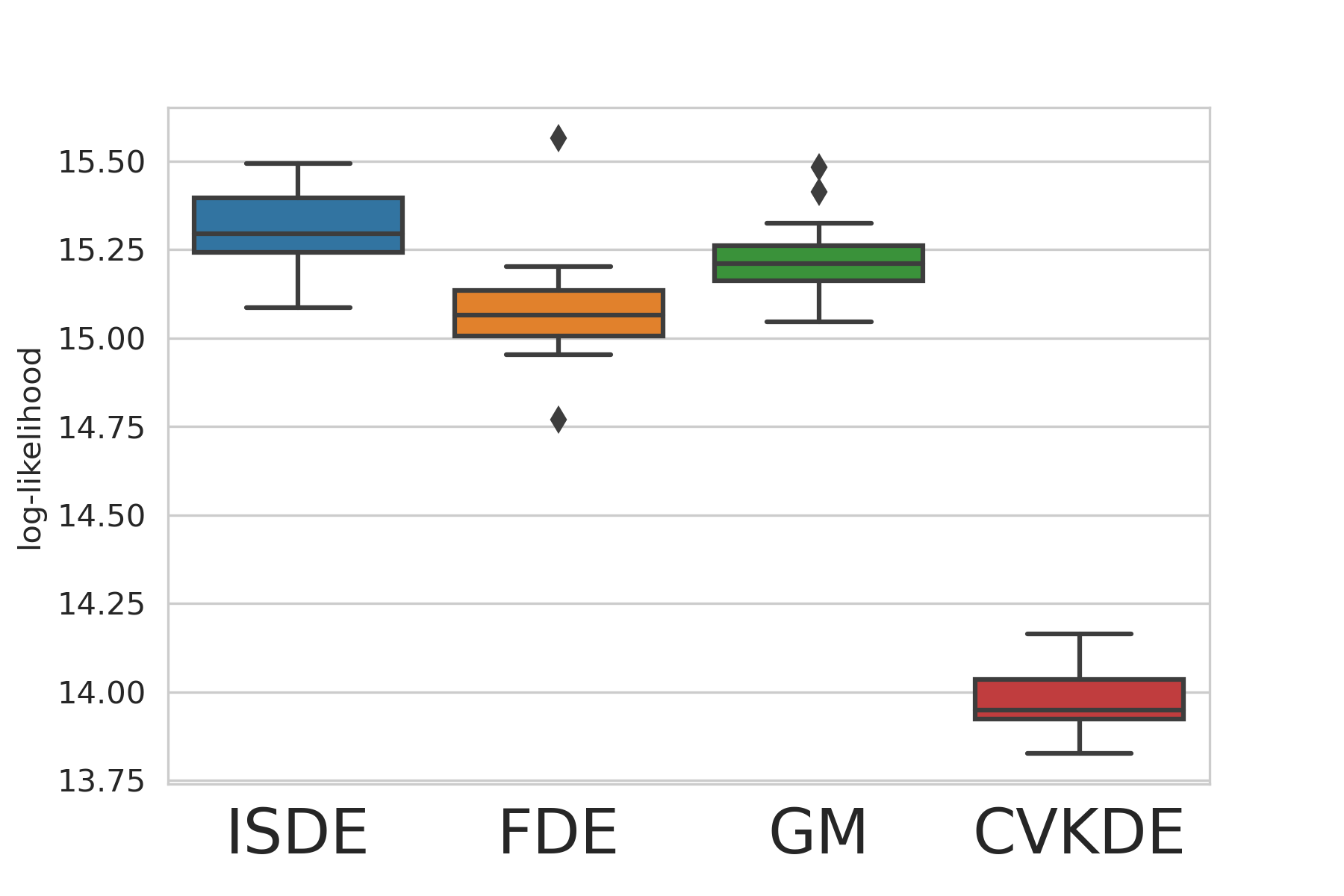} }}%
    \qquad
    \subfloat[\centering Levine32]{{\includegraphics[width=5cm]{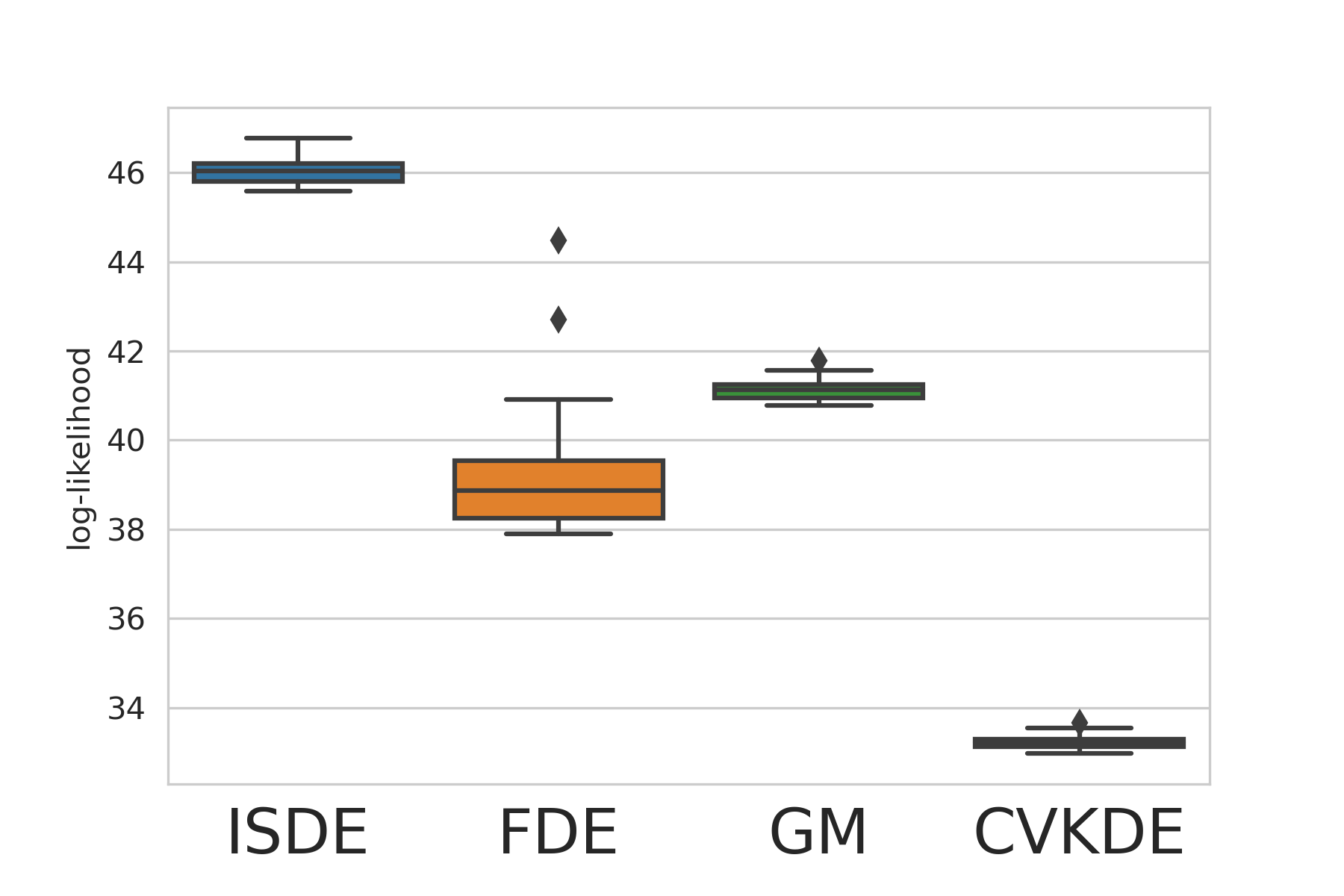} }}%
    \caption{Comparison of empirical $\log$-likelihood on validation data for different density estimation methods}%
    \label{fig:comparisonlevine}%
\end{figure}

\paragraph{Results} Boxplots indicating the $\log$-likelihood of these estimators for validation samples can be visualized in \cref{fig:comparisonlevine}.

We remark that using ISDE leads to better empirical $\log$-likelihood on validation data. CVKDE in the ambient dimension is always the worst estimator. GM is slightly better than FDE for both datasets, and the gap between performances of FDE/GM and ISDE is higher in dimension 32 than in dimension 13. We conclude that IS with a limited size of blocks seems to be a relevant model for these datasets as ISDE could outperform other model-based approaches in terms of $\log$-likelihood.

Testing ISDE against other density estimation methods is a way to evaluate how this model can explain the data well. However, we must be careful in our conclusion. These results do not indicate that the data follow an IS, but rather that IS offers a good approximation of the data distribution.

\subsection{Qualitative Interpretation}

We believe that the added value of our method is that ISs are easy to understand and useable as a tool to interpret data. After validating the pertinence of ISDE in comparison with other methods through quantitative analysis, we now provide some insight into the capacity of ISDE to deliver meaningful qualitative information.

\paragraph{Nontriviality of Outputted Partition} The first question to ask is if the gain in terms of empirical $\log$-likelihood is due to the specific outputted partition $\Phat$ or if any other estimator $\hat{f}_\Pa$ based on a partition of features $\Pa \in \Pdk$ could achieve the same performance. To answer this question, we have computed empirical $\log$-likelihood on $10$ validation sets of size $2,000$ for the three best partitions outputted by ISDE, the three worst ones regarding the optimization task, and three random partitions in $\Pdk$. To compute not the optimal but the second one, the third one, and so on, it suffices to add constraints on the partition selection problem that artificially exclude some partitions from the optimization. To compute the worst partitions, switching the optimization from maximization to minimization suffices. Random partitions are computed by generating a random permutation $\sigma$ of $\{1, \dots, d\}$ and then gather consecutive features in $\{\sigma(1), \dots, \sigma(d)\}$ in groups with sizes drawn uniformly between $1$ and $k$.

\begin{figure}[H]%
    \centering
    \subfloat[\centering Levine13]{{\includegraphics[width=5cm]{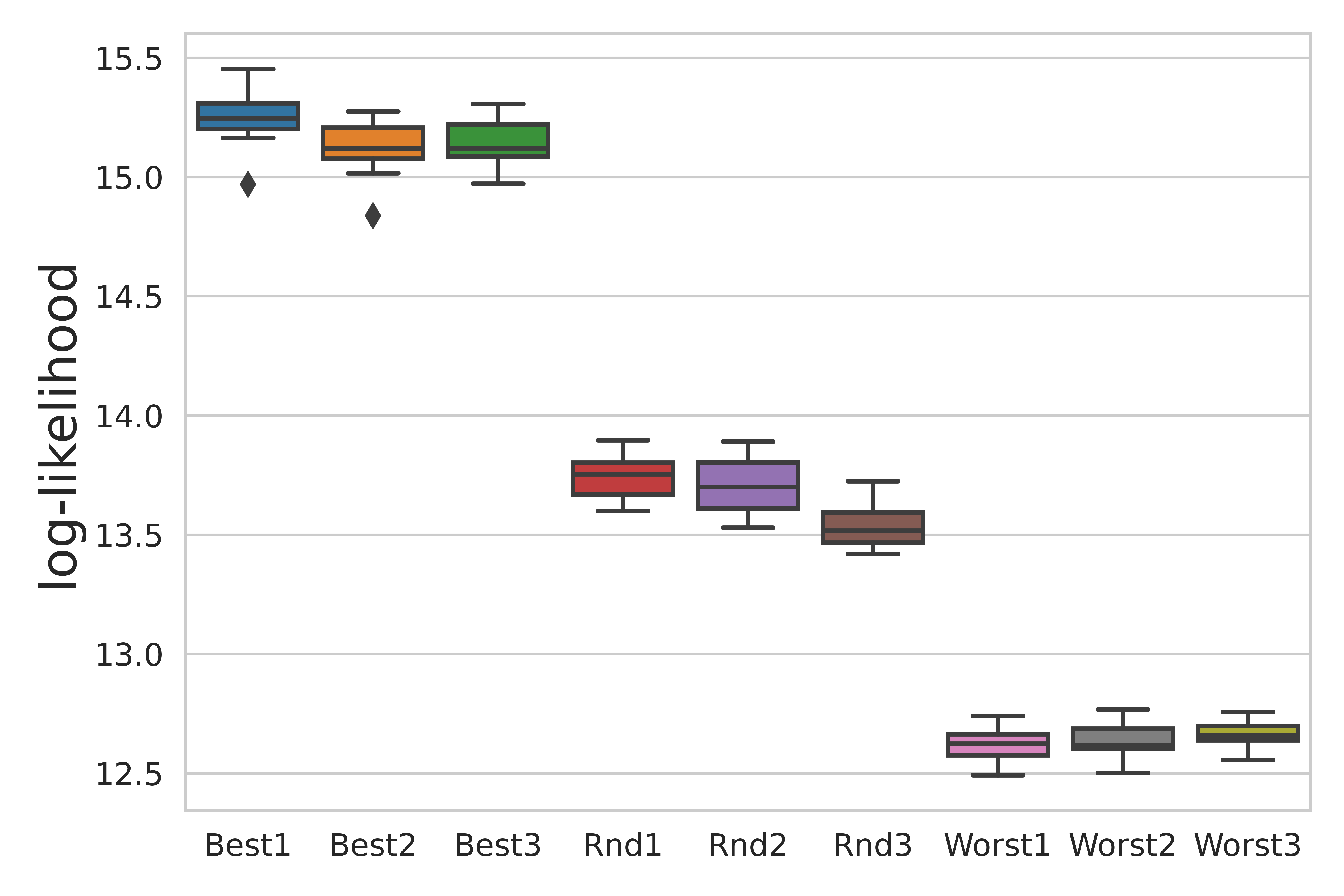} }}%
    \qquad
    \subfloat[\centering Levine32]{{\includegraphics[width=5cm]{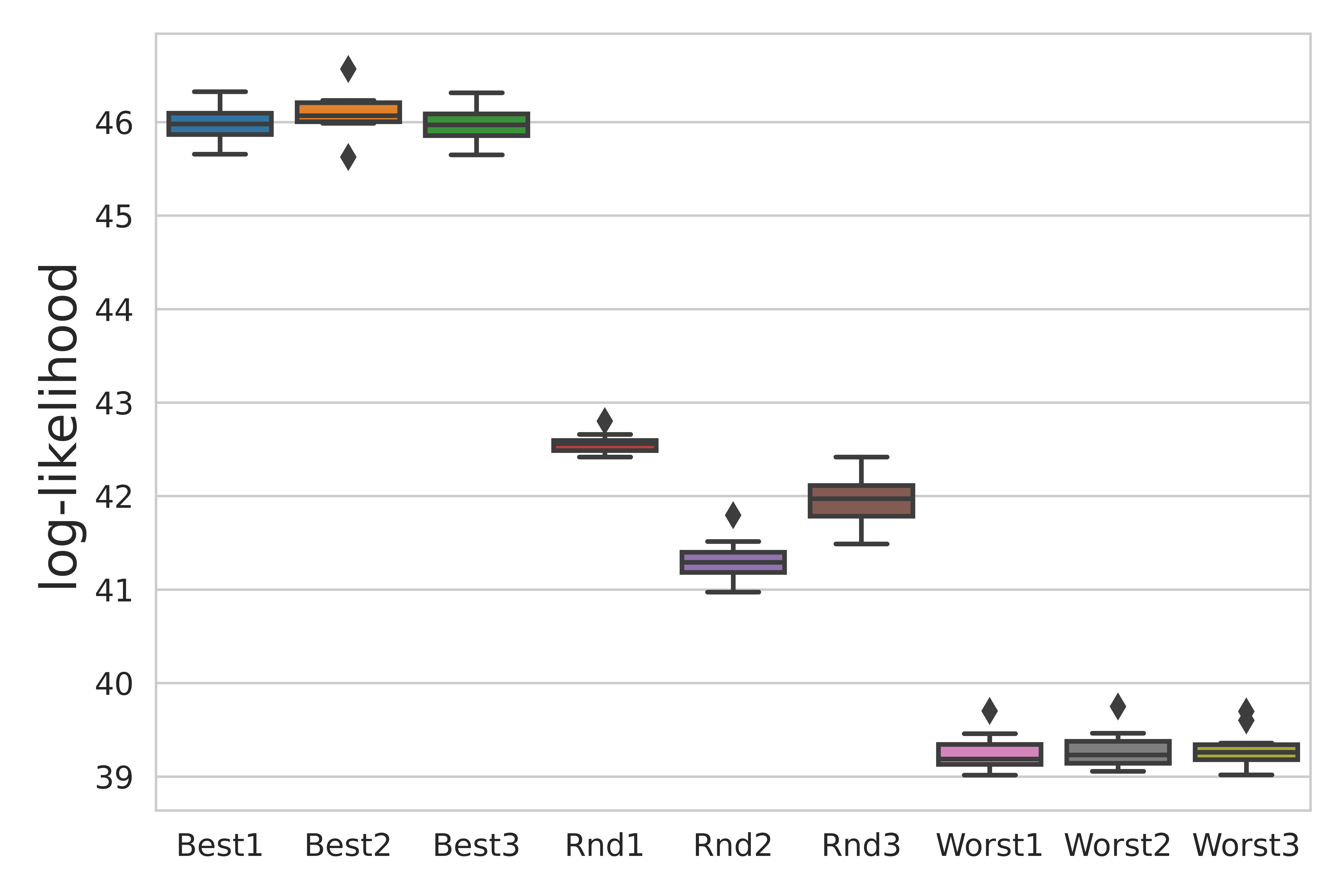} }}%
    \caption{Comparison of empirical $\log$-likelihood on validation data for best, worst and random partitions}%
    \label{fig:bestrndwosrt}%
\end{figure}

These experiments indicate that ISDE outputs specific partitions that lead to better estimators in terms of $\log$-likelihood on empirical data than the random partitions. In that sense, the information provided by ISDE on these datasets is not trivial. It also seems that not only the optimal one $\Phat$ but a collection of partitions lead to the best scores.

With that in mind, it could be interesting to determine if the collection of partitions leading to optimal results are close in some sense. To this end, it is necessary to introduce a notion of distance between partitions.

\paragraph{Edit Distance} Given two partition $\Pa$ and $\Pa'$ in $\Pdk$ it is possible to define a distance between $\Pa$ and $\Pa'$ called edit distance (\cite{brown2007automated}) and denoted by $\text{edit}(\Pa, \Pa')$. This distance corresponds to the minimal number of operations required to go from $\Pa$ to $\Pa'$ where an operation can split a block into two or merge two blocks. The edit distance defines a distance on $\Pdk$ in the mathematical sense as it is nonnegative, symmetric, equal to zero only if we compute the distance from one partition to itself and it satisfies the triangular inequality.

\paragraph{Correlation between Edit Distance and Density Estimation} We will now see how the edit distance from $\Phat$ to $\Pa$ correlates with the empirical $\log$-likelihood on validation data for $\hat{f}_\Pa$.

Firstly, we can visualize the edit distance from $\Phat$ to the $10$ best partitions (excluding $\Phat$) in the sense of the problem of partition selection, $10$ random partitions, and the $10$ worst partitions.

\begin{figure}[H]%
    \centering
    \subfloat[\centering Levine13]{{\includegraphics[width=5cm]{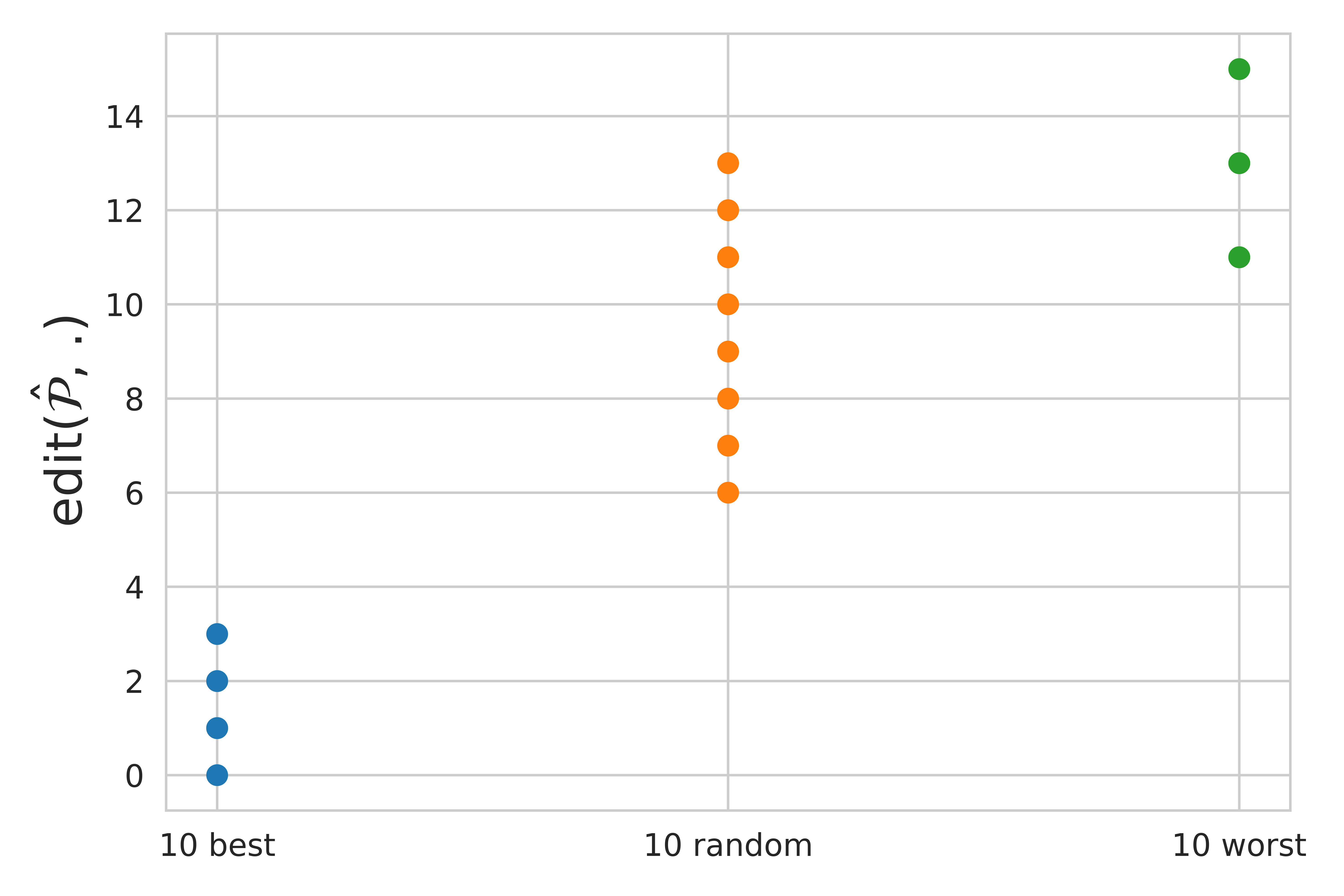} }}%
    \qquad
    \subfloat[\centering Levine32]{{\includegraphics[width=5cm]{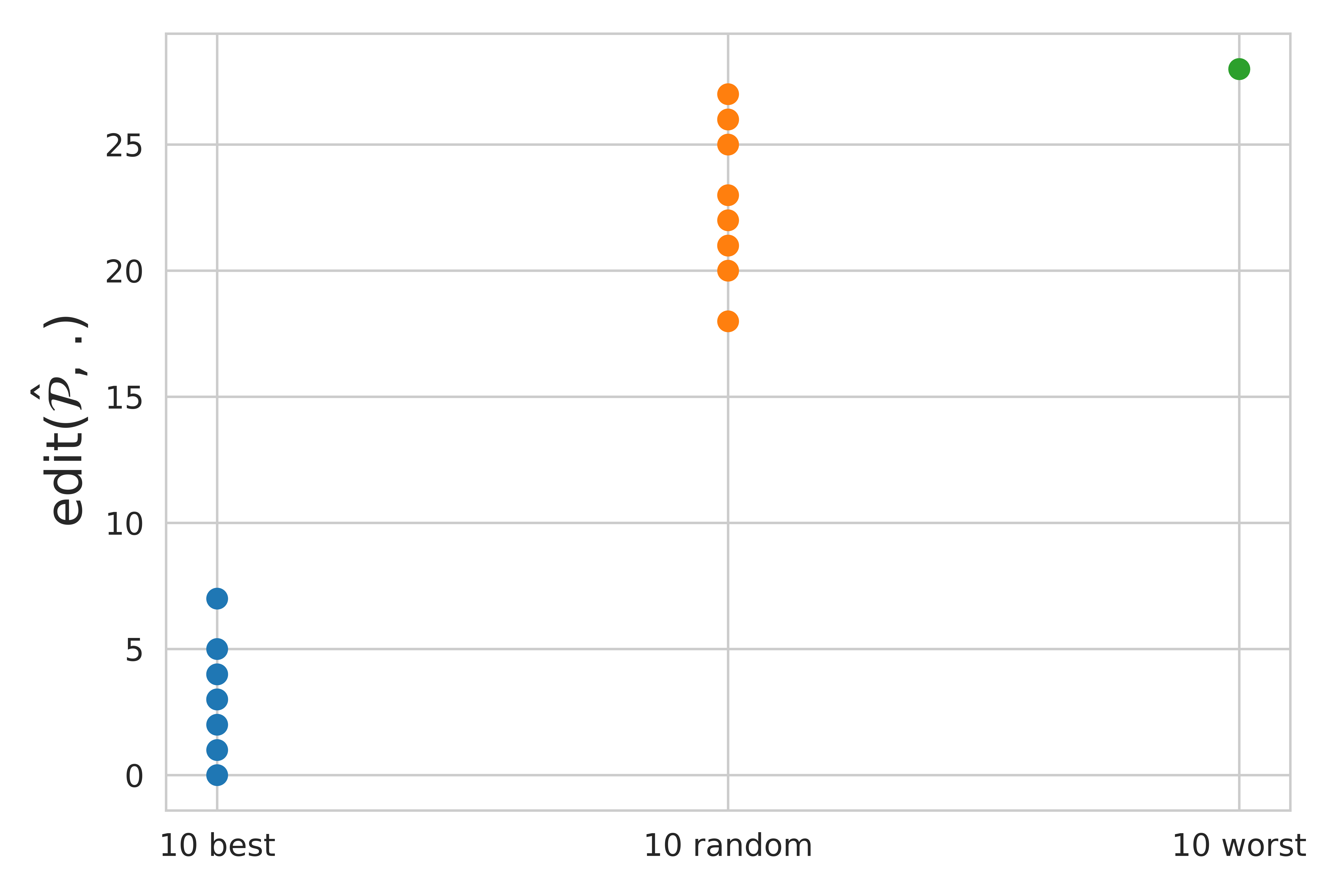} }}%
    \caption{Edit distance from $\Phat$ for $10$ best, $10$ random and $10$ worst partitions}%
    \label{fig:edit_bestrndwosrt}%
\end{figure}

These observations seem to correlate well with what we have observed previously in terms of $\log$-likelihood.

Secondly, we explore the space $\Pdk$ by defining a random walk considering the topology induced by edit. We define a random walk $(\Pa_0, \Pa_1, \dots)$ as follows: at each step we go from $\Pa_i$ to $\Pa_{i+1}$ with $\text{edit}(\Pa_i, \Pa_{i+1}) = 1$. To do so, it suffices to randomly choose an operation (edit or merge) and apply it to randomly selected block(s) of $\Pa_i$ while controlling that we stay in $\Pdk$.

To observe a possible correlation between $\text{edit}(\Phat, .)$ and $\log$-likelihood on validation data, we have implemented the following protocol: do $5$ walks of length $40$ with $\Phat$ as starting point and store all visited partitions, then for the $200$ selected partitions, compute empirical $\log$-likelihood on ten resampling of validation data and store the mean value. Then we plot these scores against $\text{edit}(\Phat, .)$.

\begin{figure}[H]%
    \centering
    \subfloat[\centering Levine13]{{\includegraphics[width=5cm]{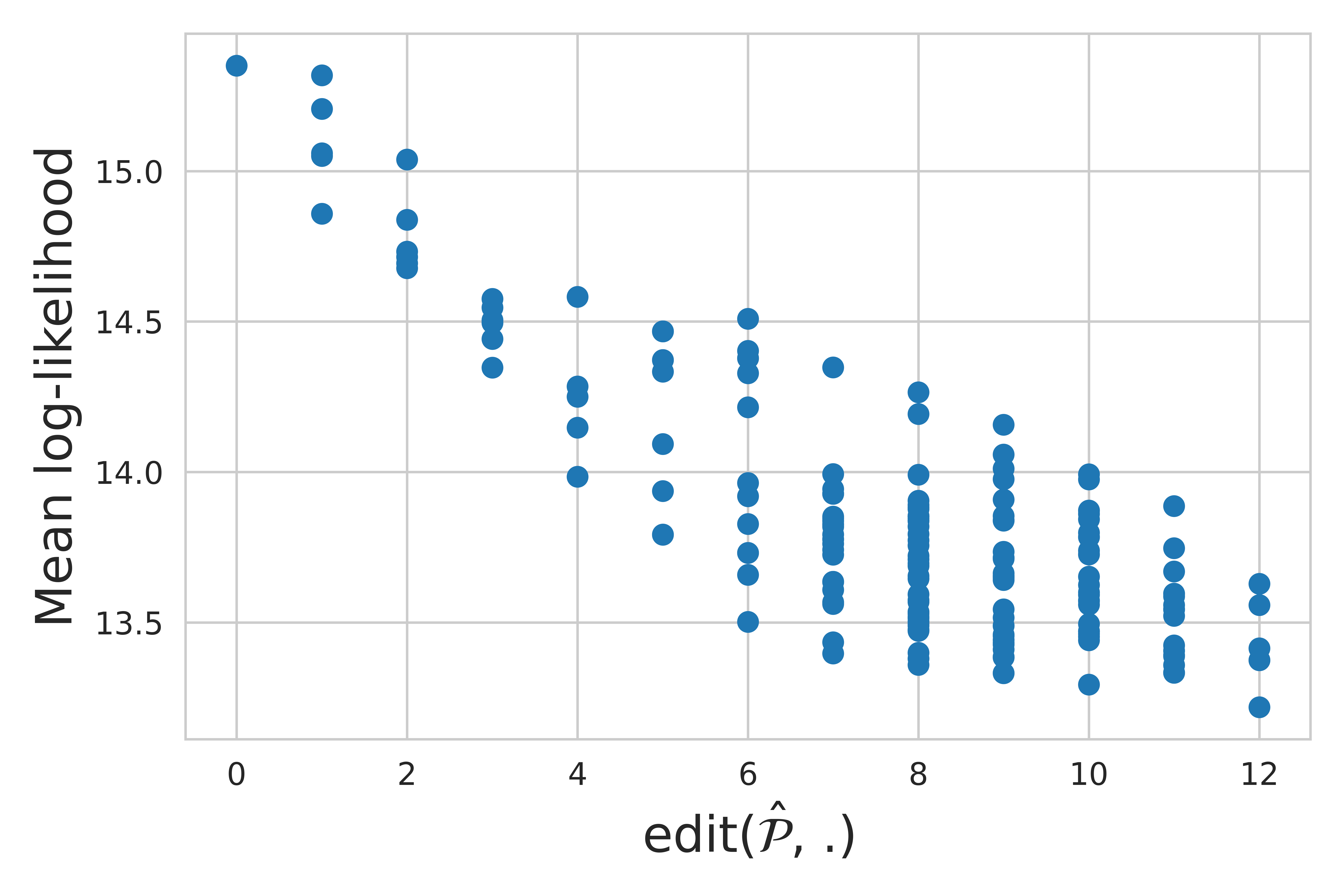} }}%
    \qquad
    \subfloat[\centering Levine32]{{\includegraphics[width=5cm]{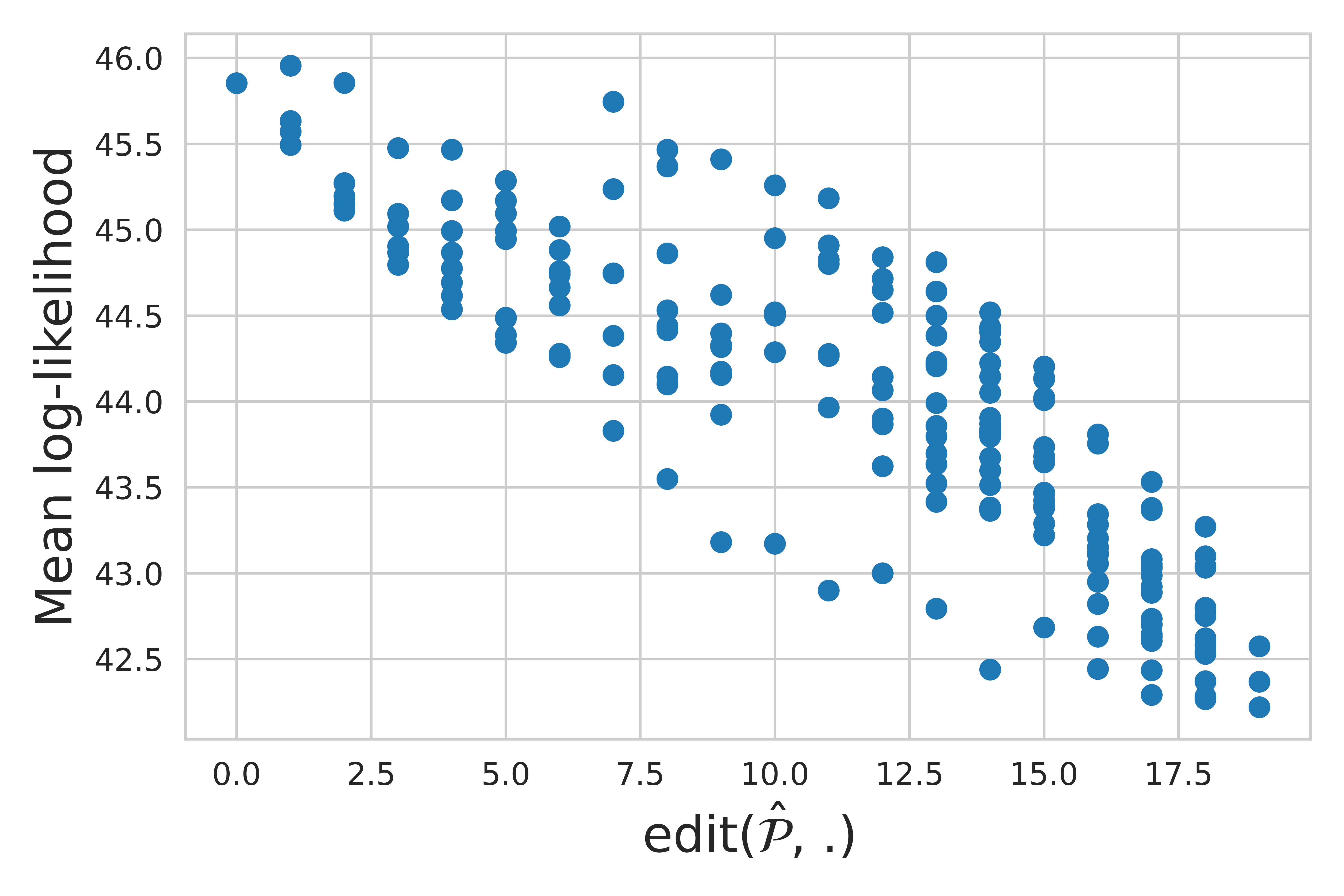} }}%
    \caption{Mean $\log$-likelihood on validation data with respect to edit distance from $\Phat$ for the partitions visited by the random walk}%
    \label{fig:ll_wrt_edit}%
\end{figure}

For both datasets, we observe a clear negative correlation between $\text{edit}(\Phat, .)$ and empirical $\log$-likelihood on validation data. These observations indicate that the topology induced by the distance edit on $\Pdk$ is meaningful in the sense that the farther a partition $\Pa$ is from $\Phat$ for the edit distance, the worse the estimator $\hat{f}_\Pa$ is.

\paragraph{Exhaustive Analysis} For the dataset Levine13, as the cardinal of $\mathrm{Part}_{13}^5$ is $25,719,630$, it is possible to store the entire family of empirical $\log$-likelihood computed thanks to the data $Z_1, \dots, Z_n$ on ISDE: $\left(\ell_n(\Pa)\right)_{\Pa \in \mathrm{Part}_{13}^5}$. Such an exhaustive analysis is impossible for Levine32 as the number of partitions in $\mathrm{Part}_{32}^3$ exceed $10^{19}$. The distribution of $\left(\ell_n(\Pa)\right)_{\Pa \in \mathrm{Part}_{13}^5}$ can be visualized thanks to an histogram.

\begin{figure}[H]%
    \centering
    \includegraphics[width=8cm]{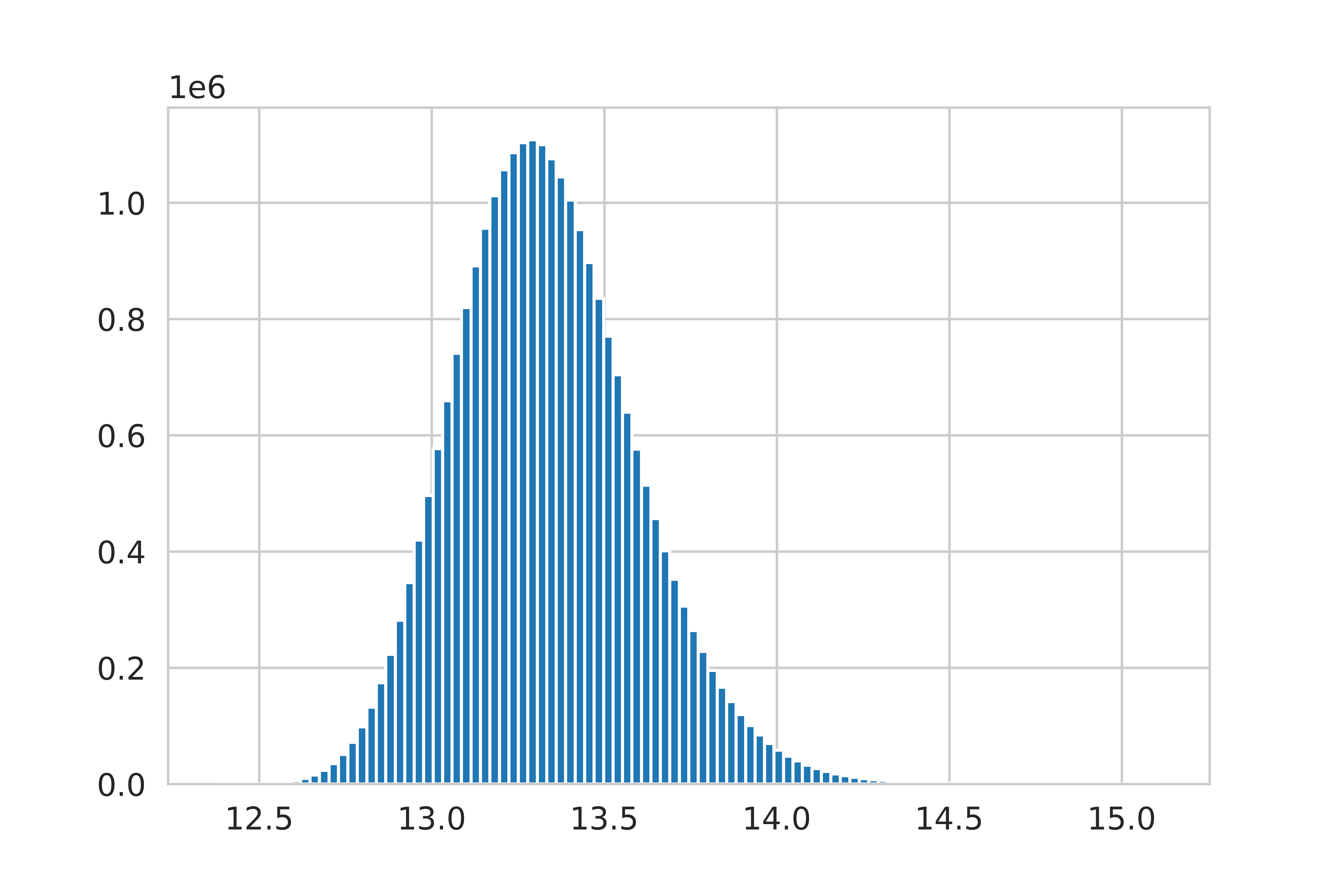}
    \caption{Distribution of $\left(\ell_n(\Pa)\right)_{\Pa \in \mathrm{Part}_{13}^5}$ }
    \label{fig:histogram_scores_levine13}%
\end{figure}

If we select the partitions with a score higher than $14.6$, there remain $1,941$ elements. For these partitions, we compute empirical $\log$-likelihood again on validation data and represent it against $\text{edit}(\Phat, .)$. This is a way to ask about the uniqueness of the optimal partition $\Phat$. If another partition $\Pa$ a significantly positive value of $\text{edit}(\Phat, \Pa)$ gives as good results as $\Phat$, it will indicate that there are other local maximums than $\Phat$.

\begin{figure}[H]%
    \centering
    \includegraphics[width=8cm]{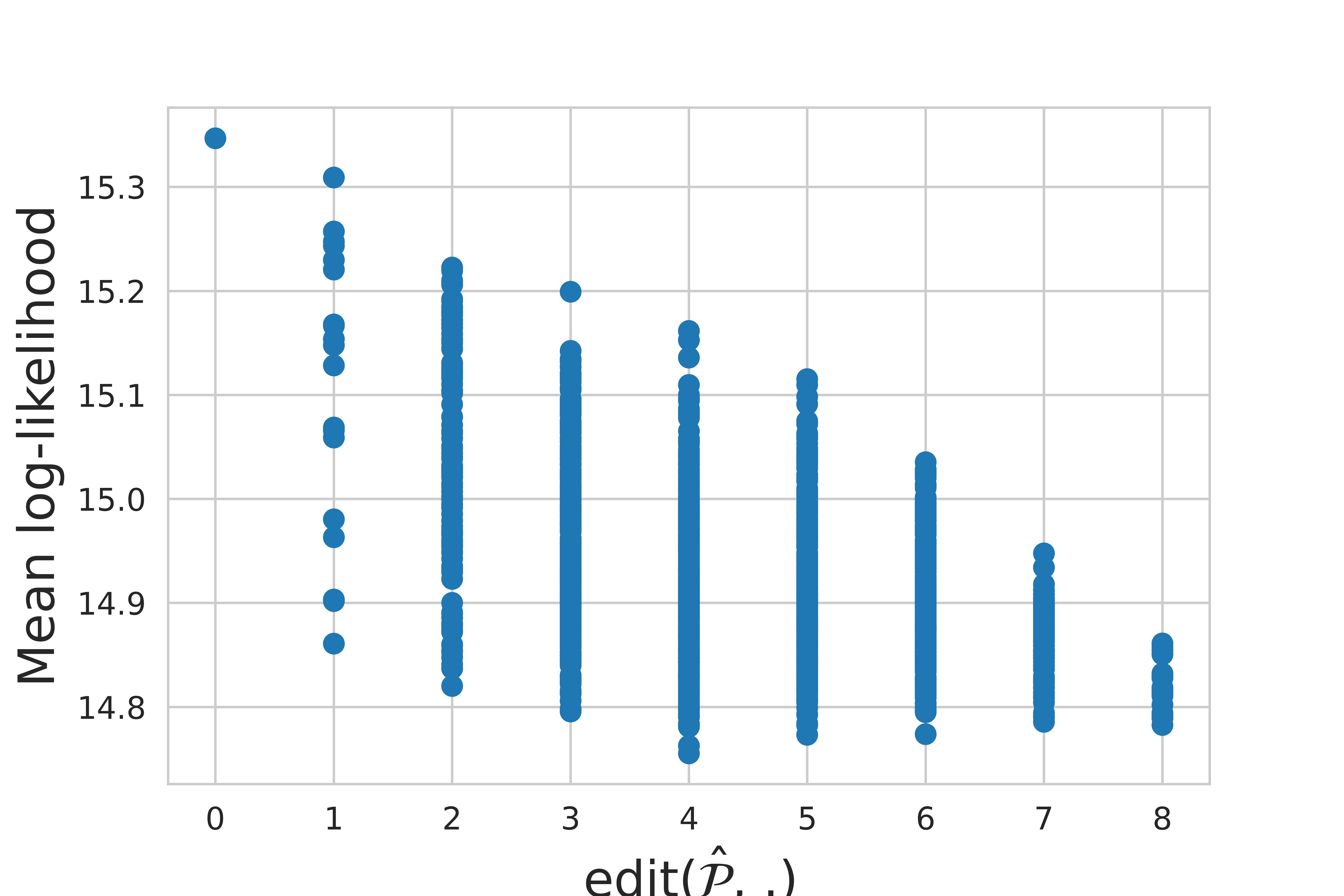}
    \caption{Mean $\log$-likelihood on validation data with respect to edit distance from $\Phat$ for $1,941$ best partitions}
    \label{fig:exhaustive_bestpartitions_levine13}%
\end{figure}

\paragraph{Conclusion} This analysis of the space $\Pdk$ equipped with edit distance in terms of empirical $\log$-likelihood for $\hat{f}_\Pa$ has led us to the conclusion that the qualitative information provided by ISDE through $\Phat$ is nontrivial for these datasets as random partitions in $\Pdk$ does not lead to optimal scores. We also show that the density estimation score deteriorates as the edit distance from $\Phat$ increases, indicating that edit distance is a relevant metric to explore $\Pdk$ in density estimation under IS. Then an exhaustive analysis of the space of partitions for Levine13 indicates that we can consider the optimal partition as unique for this experiment.

These conclusions depend on the specific datasets presented here and could become invalid for other ones. We provide the code to reproduce our experiments. Our aim is that anyone interested in the method can replicate these analyses for other data.

\section{COMPLEXITY AND RUNNING TIME ANALYSIS}\label{sec:runningtime}

In this section, we provide information about the algorithmic complexity and running time of ISDE.

\paragraph{Computation of KDE} For a given bandwidth $h$, the evaluation of a KDE constructed over $m_1$ points and evaluated over $m_2$ points is $O(m_1 m_2)$. The family of estimators $\left( \hat{f}_S \right)_{S \in \Sdk}$ is constructed using a $V$-fold cross-validation where $V$ is a divisor of $m$. If $n_h$ denotes the number of candidate values for the bandwidths, the number of operation required for bandwidth selection is $S_d^k n_h V \frac{m}{V} \times \frac{m (V-1)}{V}$. The complexity of this step is $O(S_d^k n_h m^2)$. Once the bandwidths are selected, it remains to compute the quantities $\left( \ell_n(S) \right)_{S \in \Sdk}$ thanks to $Z_1, \dots, Z_n$. The total cost of its operation is $O(S_d^k n m)$. The total algorithmic cost of the computation of $\left( \ell_n(S) \right)_{S \in \Sdk}$ is
\begin{equation}
    O\left(S_d^k m \left( n_h m + n \right) \right).
\end{equation}

\paragraph{Partition Selection} The implementation of the partition selection step relies on the branch-and-bound method. It is not easy to give a precise statement about its complexity. The branch-and-bound algorithm uses a tree search strategy to enumerate all possible solutions to a given problem implicitly. A recent survey can be found in \cite{morrison2016branch}.

\paragraph{Running time} We now present some information about running time. We have run all experiments on a laptop with the following hardware: CPU Intel(R) Xeon(R) W-10885M CPU @ 2.40GHz and GPU: Nvidia Quadro RTX 3000 Mobile.

The KDE computations have been performed on GPU using the python package pyKeOps \cite{JMLR:v22:20-275}. This implementation is much faster than the one on CPU proposed by scikit learn as highlighted by \cref{runningtimeKDE}, which compares running time for KDE constructed on $n$ points and evaluated on $n$ points on dimension $d=3$.

\begin{table}[H]
\centering
\resizebox{380pt}{!}{%
\begin{tabular}{|c|c|c|c|c|c|c|}
\hline
n & $100$ & $500$ & $2000$ & $5,000$ & $10,000$ & $20,000$ \\
\hline
\hline
GPU-based implementation & $0.0006$ & $0.0020$ & $0.0073$ & $0.0176$ & $0.0684$ & $0.1163$ \\ 
\hline
Scikit-learn implementation & $0.0008$ & $0.0126$ &  $0.1952$ & $1.1564$ &  $4.9151$ & $21.3631$ \\
\hline
\end{tabular}%
}
\caption{Comparison of running time (seconds) of sklearn implementation and ours for KDE constructed on $n$ points and evaluated on $n$ points}
\label{runningtimeKDE}
\end{table}

The computation of the quantities $\left( \ell_n(S) \right)_{S \in \Sdk}$ requires many repetitions of KDE evaluation. In \cref{totalrunningtime} we provide estimation of the running time for this step for various values of $k$ and $d$ and considering a $5$-fold cross-validation to estimate each bandwidth among $30$ candidate values. The quantities $m$ and $n$ are both set to $1,000$.

\begin{table}[H]
\centering
\resizebox{300pt}{!}{%
\begin{tabular}{|c|c|c|c|c|c|c|}
\hline
\backslashbox{k}{d}  & $5$ & $10$ & $20$ & $30$ & $40$ & $50$ \\
\hline
\hline
$2$ & $2.0$ & $6.8$ & $18$ & $40$ & $69.15$ & $108$ \\
\hline
$3$ & $3.3$ & $23$ & $121$ & $409$ & $949$ & $1,862$ \\
\hline
$4$ & $3.9$ & $53$ & $590$ & $3,053$ & $9,572$ & $23,536$ \\
\hline
$5$ & $4.2$ & $61$ & $2,154$ & $17,589$ & $75,228$ & $233,323$ \\
\hline
\end{tabular}%
}
\caption{Running time (seconds) for $\left( \ell_n(S) \right)_{S \in \Sdk}$ computation with respect to $k$ and $d$ and with $5$-fold cross selected bandwidths over $30$ possible values and for $m=n=1000$}
\label{totalrunningtime}
\end{table}

Once the quantities $\left( \ell_n(S) \right)_{S \in \Sdk}$ are computed, it remains to perform partition selection. As mentioned before we use the python package Pulp \cite{Mitchell11pulp:a}. The running time of this step for different values of $k$ and $d$ are presented in \cref{partselectionrunningtime}.

\begin{table}[H]
\centering
\begin{tabular}{|c|c|c|c|c|c|c|}
\hline
\backslashbox{k}{d}  & $5$ & $10$ & $20$ & $30$ & $40$ & $50$ \\
\hline
\hline
$2$ & $0.02$ & $0.03$ & $0.08$ & $0.20$ & $0.47$ & $0.84$ \\
\hline
$3$ & $0.02$ & $0.05$ & $0.41$ & $1.9$ & $6.2$ & $15$ \\
\hline
$4$ & $0.02$ & $0.09$ & $2.0$ & $14.8$ & $62$ & $190$ \\
\hline
$5$ & $0.02$ & $0.13$ & $7.3$ & $84.7$ & $482$ & $2,045$ \\
\hline
\end{tabular}%
\caption{Running time (seconds) for partition selection step with respect to $k$ and $d$}
\label{partselectionrunningtime}
\end{table}

The main conclusion of this running time study is that the running time of partition selection is negligible in comparison with the one for computing $\left( \ell_n(S) \right)_{S \in \Sdk}$ for the parameters presented here. The code associated with this paper contains functions allowing the reader to reproduce these experiments with different settings and estimate the running time on its device. Note that the code also runs if no GPU is available. In this case, pyKeOps will automatically use parallelization on CPU for KDE evaluations.

\section{CONCLUSION}\label{sec:conclusion}

ISDE is an algorithm that outputs an estimate of a density function of a point cloud, taking into account an IS for data in moderately high dimensions. To design it, we reduced the number of hyperparameters with an appropriate choice of the loss function and, through linear programming reformulation, made the partition selection step faster than was previously possible. This leads to reasonable running time even on a laptop for the considered datasets. The code is available and ready to be used by anyone interested in this method.

ISDE is versatile: it takes any basic multidimensional density estimator as input. Then it can be used in parametric and nonparametric frameworks. It is also exhaustive as it searches over all partitions of features with given maximal block size. To our knowledge, we are the first to propose a method that considers IS in the context of nonparametric density estimation with KDE.

We validated its performance on synthetic data satisfying IS. This performance was measured in terms of $\log$-likelihood on the validation sample. We found that ISDE exploits IS structure and outperform other density estimators for this task. Applying ISDE to mass cytometry data has indicated that it could accurately estimate density over real-world datasets and extract qualitative information about their features through the outputted partition.

This paper focused on algorithmic and implementation details relative to ISDE and empirical study. Theoretical study of ISDE will be presented in a separate work, as it involves some minor modifications to prove convergence rates.

\paragraph{Code availability} The code to reproduce the experiments presented here is available at \href{https://github.com/Louis-Pujol/ISDE-Paper}{https://github.com/Louis-Pujol/ISDE-Paper}.

\paragraph{Data availability} Original datasets were downloaded from the repository presented in \cite{weber2016comparison} and available at the address \\ \href{https://flowrepository.org/id/FR-FCM-ZZPH}{https://flowrepository.org/id/FR-FCM-ZZPH}.

\paragraph{Acknowledgement} This work was supported by the program \href{https://www.iledefrance.fr/paris-region-phd-2021}{Paris Region Ph.D.} of \href{https://www.dim-mathinnov.fr/}{DIM Mathinnov} and was partly supported by the French ANR Chair in Artificial Intelligence TopAI - ANR-19-CHIA-0001. The author is thankful to Marc Glisse and Pascal Massart for their constructive remarks on this work.

\bibliographystyle{plain}
\bibliography{bib}

\appendix
\section{APPENDIX: EXPERIMENTS ON GAUSSIAN SYNTHETIC DATA}

This section is dedicated to the presentation of synthetic results, in the same spirit as \cref{sec:synthetic} but with data drawn from centered multivariate Gaussian distributions.

\paragraph{Data Generating Process} The Gaussian Graphical Models (GGM) theory indicates that edges of the undirected graphical model associated with a Gaussian distribution $\mathcal{N}(0, \Sigma)$ are the non-zero entries of the precision matrix $\Sigma^{-1}$. As the inverse operator preserves the block-diagonal structure, we can easily simulate data from a multivariate Gaussian with an IS.

For a positive integer $s$ and a real number $\sigma \in (0, 1)$ we denotes by $\Sigma^s_\sigma$ the $s \times s$ matrix whose diagonal entries are $1$ and nondiagonal entries are $\sigma$. Then for a list of positive integers $S = [s_1, \dots s_K]$ we define the block diagonal matrix:

\begin{equation}
    \Sigma_\sigma^S = 
    \begin{pmatrix}
      \Sigma_\sigma^{s_1}
      & \rvline & \bigzero & \rvline & \dots & \rvline & \bigzero \\
    \hline
      \bigzero & \rvline &
      \Sigma_\sigma^{s_2}
      & \rvline & \ddots & \rvline & \vdots \\
      \hline
      \vdots & \rvline & \ddots & \rvline & \ddots & \rvline & \bigzero \\
      \hline
      \bigzero & \rvline & \dots & \rvline  & \bigzero & \rvline & \Sigma_\sigma^{s_K}
    \end{pmatrix}
\end{equation}

The distribution $\mathcal{N}\left(0, \Sigma_\sigma^S \right)$ satisfies the IS condition with partition \\ $\left( \left\{ \sum_{i = 1}^{j-1} s_i + 1, \dots, \sum_{i=1}^j s_i \right\} \right)_{j=1, \dots, K}$.

\paragraph{Evaluation Scheme} If $\hat{\Sigma}$ and $\Sigma$ are respectively the estimated and the true covariance, the Kullback-Leibler risk can be explicitly computed (see \cref{computKLgaussian}):

\begin{equation}
    \KL{\mathcal{N}(0, \Sigma)}{\mathcal{N}(0, \hat{\Sigma})} = \sum_{v \in \mathrm{Sp}(A) } \frac{v - \log(1 + v)}{2}
\end{equation}
where $A = (\hat{\Sigma}^{-1} - \Sigma^{-1}) \Sigma$.

\paragraph{Benchmarked Methods} Two methods will be compared to ISDE for the task of covariance estimation.

The first estimator is the simple \textbf{Empirical Covariance}, which is the maximum likelihood estimator if the covariance does not enjoy any particular structure.

The second estimator is \textbf{Block-Diagonal Covariance Selection} (BDCS) developed in \cite{devijver2018block}. It aims to estimate an IS in the context of GGM. This algorithm works in two steps:
\begin{itemize}
    \item Compute a family of nested partitions candidates to be the IS
    \item Choose a partition in this family using a slope heuristic approach
\end{itemize}

More details can be found in the original paper. Up to our knowledge, this is the only work dealing specifically with IS in the GGM framework.
 
\paragraph{ISDE Inputs} We run \cref{algo} with $k = d$, $m = n = 0.5 \times N$ and simple empirical covariance as multivariate density estimator.

\paragraph{Performance} We compare the three methods described above for fixed $\sigma$, $N$, and different structures $S$. We have gathered results in terms of KL loss are in \cref{SvaryGaussian}. We have repeated each experiment $5$ times, and the scores displayed are the mean KL losses and standard deviation over these repetitions.

\begin{table}[H]
\centering
\resizebox{330pt}{!}{%
\begin{tabular}{|l|c|c|c|c|}
\hline
 S & [2, 2] & [4, 4, 1] & [4, 3, 2, 3] & [4, 4, 3, 3 ,2 ] \\
\hline
\hline
ISDE & $\mathbf{0.60 \pm 0.21}$ & $1.88 \pm 0.52$ & $2.85 \pm 0.60$ & $5.30 \pm 0.96$\\
\hline
BDCS & $\mathbf{0.60 \pm 0.21}$ & $\mathbf{1.72 \pm 0.46}$ & $\mathbf{2.63 \pm 1.01}$ & $\mathbf{4.42 \pm 1.80}$ \\
\hline
Empirical & $0.80 \pm 0.20$ & $3.62 \pm 0.53$ & $6.88 \pm 0.84$ & $12.63 \pm 0.83$\\
\hline
\end{tabular}%
}
\caption{Gaussian: KL Losses ($. 10^3$) - $\sigma = 0.7$, $N = 6000$} \label{SvaryGaussian}
\end{table}

\paragraph{Recovery} We are interested not only in performance, but we also want to find the correct partition in order to get qualitative information about datasets. In \cref{recoveryGaussian} we collect, for the same experiment as above, the rate of recovery of the proper partition. In parentheses is displayed the rate of admissible output partition: a partition is admissible if all the blocks of the original partition are subsets of blocks of this one.

\begin{table}[H]
\centering
\resizebox{330pt}{!}{%
\begin{tabular}{|l|c|c|c|c|}
\hline
S & [2, 2] & [4, 4, 1] & [4, 3, 2, 3] & [4, 4, 3, 3 ,2 ] \\
\hline
\hline
ISDE& 100\%(100\%) & 80\%(100\%) & 40\%(100\%) & 0\%(100\%) \\
\hline
BDCS & 100\%(100\%) & 100\%(100\%) & 80\%(100\%) & 60\%(100\%) \\
\hline
\end{tabular}%
}
\caption{Gaussian: Recovery - $\sigma = 0.7$, $N = 6000$} \label{recoveryGaussian}
\end{table}

\paragraph{Conclusion} We remark that BDCS is the most efficient method for the task of density estimation in GGM under IS. We can explain it as ISDE tends to select admissible partition but fails to select the exact IS when the dimension grows. BDCS inherently penalizes more useless blocks merging, making it more accurate in this setting.

However, ISDE performs significantly better than a naive empirical covariance, proving that it benefits from the IS.

We want to highlight the difference between ISDE and BDCS. BDCS starts by selecting a family of up to $d$ nested partitions and then selects among them. This approach uses a preliminary covariance estimator to design this family of nested partitions. This approach is reasonable as for Gaussian data, pairwise dependencies entirely determine multidimensional dependencies between features. Outside the scope of GGM, this approach does not remain valid as features of a random variable can be pairwise independent but mutually dependent. ISDE can handle more general settings as it selects among a set of partitions with blocks of cardinal potentially more significant than 2.

\section{APPENDIX: TECHNICAL RESULTS}

\subsection{Computation of $B_d^2$}\label{computB2d}

Let us prove the following formula :
\begin{align}
    B_d^2 &= \sum_{i=1}^{\lfloor d / 2 \rfloor} \frac{\prod_{j=0}^{i-1} \binom{d-2j}{2}}{i !} \\
    &= 1 + \binom{d}{2} + \frac{\binom{d}{2}\binom{d-2}{2}}{2!} + \frac{\binom{d}{2}\binom{d-2}{2}\binom{d-4}{2}}{3!} \dots + \frac{\binom{d}{2} \dots \binom{d - 2\left(\lfloor d/2 \rfloor - 1\right)}{2}}{(\lfloor d / 2\rfloor )!}
\end{align}

For a nonnegative integer $i$, let us denote by $B_d^2[i]$ the number of partitions of $\Pdk$ with exactly $i$ blocks of size $2$. A first remark is that $B_d^2[i] = 0$ as soon as $i > \lfloor d / 2 \rfloor$, then
\begin{equation}
    B_d^2 = \sum_{i=0}^{\lfloor d / 2 \rfloor} B_d^2[i].
\end{equation}

Now, we evaluate $B_d^2[i]$. It is not hard to count the number of possibilities to select $i$ pairs of distinct elements of $\{1, \dots, d\}$ taking into account in which order there were selected. For the first pair, there are $\binom{d}{2}$ choices, then $\binom{d-2}{2}$ choices for selecting another pair among the other variables, and so on. Then there are $\prod_{j=0}^{i-1} \binom{d-2j}{2}$ ordered pairs of variables of $\{1, \dots, d\}$.

As selecting a partition in $\Pdk$ is equivalent to an unordered choice of pairs of variables, it remains to divide by the number of permutation of $i$ elements, $i!$. Then
\begin{equation}
    B_d^2[i] = \frac{\prod_{j=0}^{i-1} \binom{d-2j}{2}}{i!}.
\end{equation}

\subsection{Computation of $\KL{\mathcal{N}(0, {\Sigma_1})}{\mathcal{N}(0, \Sigma_2)}$}\label{computKLgaussian}

Let us prove that if ${\Sigma_1}$ and $\Sigma_2$ are two covariance matrix, then
\begin{equation}
    \KL{\mathcal{N}(0, {\Sigma_1})}{\mathcal{N}(0, \Sigma_2)} = \sum_{v \in \mathrm{Sp}(A) } \frac{v - \log(1 + v)}{2}
\end{equation}
where $A = (\Sigma_2^{-1} - {\Sigma_1}^{-1}) {\Sigma_1}$.

First of all, for a covariance matrix $\Sigma$, the density $f_\Sigma$ of $\mathcal{N}(0, \Sigma)$ is given by
\begin{equation}
    \forall x \in \mathbb{R}^d f_\Sigma(x) = \frac{1}{(2 \pi)^{d / 2} (\det \Sigma)^{1 / 2}} \exp\left( - \frac{1}{2} x^\mathrm{T}\Sigma^{-1}x \right).
\end{equation}

We compute the KL divergence between $f_{\Sigma_1}$ and $f_{\Sigma_2}$

\begin{align}
    \KL{f_{\Sigma_1}}{f_{\Sigma_2}} &= \int \log\left( \frac{f_{\Sigma_1}(x)}{f_{\Sigma_2}(x)} \right) f_{\Sigma_1}(x) dx \\
    &= \frac{1}{2} \log\frac{\det \Sigma_2}{\det {\Sigma_1}} \underbrace{\int f_{\Sigma_1}(x)dx}_{=1} \\
    &\ \ +\frac{1}{2} \underbrace{\int x^\mathrm{T}\Sigma_2^{-1}x f_{\Sigma_1}(x)dx}_{= \mathrm{Tr}(\Sigma_2^{-1}{\Sigma_1})} \\
    &\ \ + \frac{1}{2} \underbrace{\int x^\mathrm{T}{\Sigma_1}^{-1}x f_{\Sigma_1}(x)dx}_{= \mathrm{Tr}({\Sigma_1}^{-1}{\Sigma_1}) = d} \\
    &= \frac{1}{2}\left( \log \det \Sigma_2 - \log \det {\Sigma_1} + \mathrm{Tr}\left( \Sigma_2^{-1} {\Sigma_1} \right) - d \right) \label{KLgaussianMatrix}
\end{align}

We remark that
\begin{equation}
    \mathrm{Tr}\left( \Sigma_2^{-1} {\Sigma_1} \right) - d = \mathrm{Tr}(A) = \sum_{v \in \mathrm{Sp}(A)} v
\end{equation}

We also remark that $\log \left( \frac{\det {\Sigma_1}}{\det \Sigma_2} \right) = \log\left( \det \Sigma_2^{-1}{\Sigma_1} \right)$ and as if $v$ is an eigenvalue of $A$, $1+v$ is an eigenvalue of $\Sigma_2^{-1}{\Sigma_1}$ we have
\begin{equation}
    \log \left( \frac{\det {\Sigma_1}}{\det \Sigma_2} \right) = \sum_{v \in \mathrm{Sp}(A)} \log(1 + v)
\end{equation}

Combining these results with \cref{KLgaussianMatrix} leads to the desired formula.

\end{document}